\documentclass[10pt, conference, compsocconf]{IEEEtran}

\usepackage{times}
\usepackage{balance}
\usepackage{setspace}
\usepackage{algorithm}
\usepackage{algorithmic}
\usepackage{boxedminipage} 
\usepackage{graphicx}
\usepackage{verbatim} 
\usepackage{amsmath}
\usepackage{times} 
\usepackage{gensymb}
\usepackage{dsfont}
\usepackage{multirow}
\usepackage{amsfonts}
\usepackage{float}
\usepackage{epstopdf}
\usepackage[T1]{fontenc}
\usepackage{url}
\usepackage{amssymb}
\usepackage[tight,footnotesize]{subfigure}
\usepackage[caption=false,font=footnotesize]{subfig}
\usepackage{epsfig}
\usepackage{bm}
\usepackage{breqn}

\begin{document}
%
\title{Analyzing different prototype selection techniques for dynamic classifier and ensemble selection}

\author{\IEEEauthorblockN{Rafael M. O. Cruz and Robert Sabourin}
\IEEEauthorblockA{\'{E}cole de technologie sup\'{e}rieure - Universit\'{e} du Qu\'{e}bec\\
Email: cruz@livia.etsmtl.ca, robert.sabourin@etsmtl.ca}
\and
\IEEEauthorblockN{George D. C. Cavalcanti}
\IEEEauthorblockA{Centro de Inform\'{a}tica - Universidade Federal de Pernambuco\\
Email: gdcc@cin.ufpe.br}
}

\maketitle

\begin{abstract}

	In dynamic selection (DS) techniques, only the most competent classifiers, for the classification of a specific test sample are selected to predict the sample's class labels. The more important step in DES techniques is estimating the competence of the base classifiers for the classification of each specific test sample. The classifiers' competence is usually estimated using the neighborhood of the test sample defined on the validation samples, called the region of competence. Thus, the performance of DS techniques is sensitive to the distribution of the validation set. In this paper, we evaluate six prototype selection techniques that work by editing the validation data in order to remove noise and redundant instances. Experiments conducted using several state-of-the-art DS techniques over 30 classification problems demonstrate that by using prototype selection techniques we can improve the classification accuracy of DS techniques and also significantly reduce the computational cost involved.

\end{abstract}


\section{Introduction}

Multiple Classifier Systems (MCS) aim to combine classifiers in order to increase the recognition accuracy in pattern recognition systems~\cite{kittler,kuncheva}. MCS are composed of three phases~\cite{Alceu2014}: (1) Generation, (2) Selection, and (3) Integration. In the first phase, a pool of classifiers is generated. In the second phase, a single classifier or a subset having the best classifiers of the pool is(are) selected. We refer to the subset of classifiers as the Ensemble of Classifiers (EoC). In the last phase, integration, the predictions of the selected classifiers are combined to obtain the final decision~\cite{kittler}.

Recent works in MCS have shown that dynamic selection (DS) techniques achieve higher classification accuracy when compared to static ones~\cite{Alceu2014,CruzPR,knora}, especially when dealing with small sized datasets~\cite{paulo2}. Dynamic selection techniques consist of, based on a pool of classifiers $C$, in finding a single classifier $c_{i}$, or an ensemble of classifiers $C'$, that has (or have) the most competent classifiers to predict the label of an unknown sample, $\mathbf{x}_{j}$. Usually, the competence of a base classifier is estimated based on instances that are similar to the query sample, using the K-Nearest Neighbors (K-NN) technique, and a set of labeled samples, which can be either the training or validation set. Such set is called the dynamic selection dataset (DSEL). The set with the K-Nearest Neighbors is called the region of competence and is denoted by $\theta_{j} = \left \{ \mathbf{x}_{1}, \ldots, \mathbf{x}_{K} \right \}$. The instances belonging to $\theta_{j}$ are used to estimate the competence level of the base classifiers, according to a given DS criteria such as local accuracy~\cite{lca}, classifier behavior~\cite{mcb,paulo2} and probabilistic models~\cite{Woloszynski,WoloszynskiKPS12}.
	
In a recent analysis\cite{reportarXiv}, we found that the performance of DS techniques is very dependent on the distribution of DSEL. When the samples in this set are not representative enough of the query sample, the DS technique may not select the most competent classifiers to predict its label. This phenomenon can happen due to the high degree of overlap between different classes or due to the presence of noise (e.g., mislabeled samples). In a recent publication~\cite{cruz2016prototype,ijcnn2011}, we show that a simple edition of the DSEL distribution, using a prototype selection (PS) technique, significantly improves the classification accuracy for several DS methods. In this case, the Edited Nearest Neighbor (ENN)~\cite{enn,T1976} was used, since it is a classical PS technique for removing noise and decreasing the amount of overlap in the class borders, producing smoother decision borders.

Another important aspect of editing the distribution of DSEL comes from the observation that, by removing samples in this set, the running time of the dynamic selection techniques decreases. For every DS technique, the running time to classify a given test instance $\mathbf{x}_{j}$ of each method is a combination of the definition of the region of competence and evaluating the competence level of each classifier in the pool. The definition of the region of competence is conducted using the KNN technique, hence, the computational cost involved is of order $O(d \times N)$, given that $d$ and $N$ are the number of dimensions and samples in the dataset. Moreover, several dynamic selection techniques pre-calculate the outputs of the base classifiers for the samples DSEL during the training stage and store them into a matrix. The storage requirement for the pre-calculated information is $O(M \times N \times \Omega)$, with $M$ and $\Omega$ being the number of classifiers in the pool and the number of classes in the dataset. Thus, by reducing the size of the dynamic selection dataset through PS techniques, we can significantly reduce the computational cost involved in applying dynamic selection techniques, making them usable for large datasets.

However, the ENN technique only removes overlapping samples, which are closer to the class borders, while redundant instances which are close to the class centers remain unchanged. As reported in~\cite{Garcia:2012}, the ENN presented one of the lowest dataset reduction rates when compared to over 50 different PS techniques. Dataset reduction rate is important for DS as the size of the dynamic selection dataset (DSEL) has the highest influence in the computational complexity of the system during the generalization steps. So, by significantly reducing the size of DSEL, such dynamic selection techniques can also be applied for large datasets. Since the ENN present a low dataset reduction rate, it does not significantly reduce the computational complexity cost involved on DS techniques. Furthermore, based on a recent comparison among PS techniques~\cite{Garcia:2012}, there are several techniques that outperform the ENN also regarding classification accuracy.

Thus, the objective of this work is to analyze the impact of different PS techniques for dynamic selection. We compare the performance of different PS techniques based on the classification accuracy as well reduction in computational time for dynamic selection techniques. A total of six PS techniques is evaluated in this work over 30 classification problems. Moreover, six state-of-the-art dynamic selection techniques are considered in this study.

In a nutshell, the following research questions are investigated in this paper:

\begin{enumerate}
	\item Is the ENN~\cite{cruz2016prototype} the best PS technique for dynamic selection techniques? 
	
	\item Do different prototype selection techniques improve the classification performance of dynamic selection techniques?

	\item Is the selection of best PS technique application dependent?
	
	\item Can we significantly reduce the computational complexity of dynamic selection techniques by reducing the size of DSEL without losing generalization performance?
\end{enumerate}

This paper is organized as follows: Section~\ref{sec:literature} presents the prototype selection techniques analyzed in this work. Experiments are conducted in Section~\ref{experiments}. The conclusion is given in the last section.

\section{Prototype selection for dynamic selection}
\label{sec:literature}

There are three types of prototype selection mechanisms available~\cite{Garcia:2012}: condensation, edition and hybrid.
\begin{itemize}
	\item Condensation techniques remove instances that are closer to the center of the classes, which are considered redundant, while the points close to the class border remain unchanged. The goal is to reduce the dataset size without losing the generalization performance of the KNN classifier.  
	
	\item Edition techniques aim to improve the performance of the KNN classifier by removing instances with a high risk of being noise. The editing process occurs in regions of the feature space with a high degree of overlap between classes, producing smoother class boundaries.
	
	\item Hybrid techniques perform both a condensation of the data and edition of the class borders.
\end{itemize}
	
Since we are interested in increasing the generalization performance of the system and reducing the computational complexity, only hybrid and edition techniques were considered. We selected the best hybrid and edition prototype selection technique which presented the best overall results according to~\cite{Garcia:2012}. 

Three prototype selection from each paradigm were evaluated. Random Mutation Hill Climbing (RMHC)~\cite{Skalak94}, Relative Neighborhood Graph (RNG)~\cite{SanchezPF97} and Edited Nearest Neighbor (ENN)~\cite{enn} as edition techniques. The following techniques were considered from the hybrid paradigm: Steady-State Memetic Algorithm (SSMA)~\cite{Garcia:2008}, CHC Evolutionary Algorithm (CHC)~\cite{Eshelman90} and, Generational Genetic Algorithm (GGA)~\cite{KunchevaJ99,Kuncheva95}. For the following definitions, we refer to the edited distribution of DSEL (i.e., the set after applying the PS method) as DSEL$^{'}$.

\subsection{Steady-State Memetic Algorithm (SSMA)}

The SSMA~\cite{Garcia:2008} is a PS technique that used evolutionary algorithm (EA), called Memetic Algorithm, that employs local searches in its procedure. The steps of the SSMA technique are summarized as follows:

\begin{enumerate}
	\item Encoding: Each chromosome is represented by a binary array. Each element of the array
	indicates if the corresponding instance of is included (1) in the solution or not included (0).
	
	\item Population initialization: A random number of instances is selected to compose the
	initial population.
	
	\item Fitness function: The fitness function in the SSMA technique balances the accuracy and the size of the dataset (Equation~\ref{eq:fitness}). Hence, this technique aims to reduce the size of the dataset and improve the classification accuracy.
		
	\begin{equation}
	\label{eq:fitness}
	Fitness = \alpha \times accuracy + (1-\alpha) \times perc_{red}
	\end{equation}
	
	\noindent where, $accuracy$ is the leave-one-out classification accuracy of the 1NN, and  $perc_{red}$ is the percentage of reduction achieved with regard to the original size of $DSEL$	
		
	\item Parent selection: A binary tournament is used to select the two parents that will generate two
	offsprings.
	
	\item Genetic operators: since the technique uses GA, two operators are used for generating the next population. For the crossover, each offspring randomly receives half of the genes from each parent. In addition, through mutation, a random gene will change its state (0 to 1 or vice versa) according to a predefined mutation probability.
	
	\item The algorithm ends when the maximum number of iterations is achieved. 
	
\end{enumerate}

After the maximum number of iterations is reached, the chromosome that achieved the best fitness is considered as the edited dynamic selection dataset, DSEL$^{'}$.

\subsection{Generational Genetic Algorithm (GGA)}

The basic idea in GGA is to maintain a population of chromosomes, which represent plausible solutions to the particular problem, and evolving this population through a process of competition and controlled variation. A binary array represents each chromosome, where each element of the array indicates whether the corresponding instance is included (1) in the dataset or not (0).


The classical GGA algorithm consists of three operations:

\begin{enumerate}
	\item Estimate the fitness of all individuals.
	\item Generation of an intermediate population (gene pool) using the selection mechanism.
	\item Combination of the selected individuals through the crossover and mutation operators.
\end{enumerate}

At each generation, the algorithm produces a new population, S, by copying the chromosomes from the previous population S(t-1). In this case, chromosomes with higher fitness S(t-1) have a greater probability of being selected to form new population S(t). The fitness of each individual is computed according to Equation~\ref{eq:fitness}. Thus, this technique aims to reduce the size of the dataset and improve the classification accuracy.

Next, the crossover and mutation operators are applied. It is important to mention that the crossover operator is not used to all pair of chromosomes in the new population. A random choice is made, in which the probability of crossover being used depends on a predefined parameter, called crossover rate. Next, the mutation operator occurs, and it consists of flipping one or more random bits in the chromosome string with a probability that is equal to the mutation rate. The algorithm ends after the maximum number of generations are reached. The chromosome that achieved the best fitness is considered as the edited dynamic selection dataset, DSEL$^{'}$.

\subsection{CHC Adaptive Search Algorithm}

Proposed in~\cite{Eshelman90}, this algorithm is also based on evolutionary computation to perform prototype selection.
Three steps are conducted at each generation of the CHC algorithm:

\begin{enumerate}
	\item It generates an intermediate population with N chromosomes.
	\item The intermediate population is paired, and used to generate N new offsprings.
	\item A survival mechanism is held, where the best N chromosomes from the intermediate population (parents) and the generated offspring are selected to form the new generation of chromosomes.
\end{enumerate}

An important point of the CHC algorithm that differs this technique from the other PS presented in this paper comes from the observation that this algorithm uses a heterogeneous recombination of the chromosomes using a half uniform crossover scheme, called HUX. This operator exchanges half of the bits that differ between parents, where the bit position to be exchanged is randomly determined.

No mutation operator is applied to increase diversity in the search by the CHC technique. However, when the search fails to improve the fitness for a consecutive number of iterations, the population is reinitialized in order to introduce more diversity to the search. The fitness of each chromosome is measured using Equation~\ref{eq:fitness}. The reinitialization occurs as follows: The chromosome that achieved the best solution (best fitness) during the search is used as a model to reinitialize N-1 chromosomes. The reinitialization process is accomplished by randomly changing 35\% of the bits in the model chromosome to generate the other N - 1 new chromosomes in the population.

After the maximum number of iterations is reached, the chromosome that achieved the best fitness is considered as the edited dynamic selection dataset, DSEL$^{'}$.

\subsection{Random Mutation Hill Climbing (RMHC)}

The RMHC method~\cite{Skalak94} works by selecting an initial subset $S$ selected randomly from DSEL. Then a random mutation is applied to $S$, which means that an instance is either added or removed from $S$ randomly. Next, the fitness of $S$ is evaluated using the accuracy of the 1NN classifier. If the resulting $S$ have a better fitness than its predecessors it is kept as the best $S_{best}$. The phases of mutation and fitness evaluation are repeated for a predefined number of iterations. The subset $S_{best}$, which achieved the best fitness is considered as the edited DSEL, DSEL$^{'}$.

\subsection{Edited Nearest Neighbor (ENN)}

Given the dynamic selection dataset DSEL, the ENN algorithm works as follows (Algorithm~\ref{alg:enn}): For each instance $\mathbf{x}_{j,DSEL} \in$ DSEL, the class label of $\mathbf{x}_{j,DSEL}$ is predicted using the KNN algorithm using a leave-one-out procedure. A K = 3 was used, as suggested by Wilson~\cite{enn}, in order to satisfy the asymptotic properties of the NN technique. If $\mathbf{x}_{j,DSEL}$ is misclassified by the KNN technique, it is removed from the set, since $\mathbf{x}_{j,DSEL}$  is located in a region of the feature space where the majority of samples belongs to a different class. The edited dynamic selection dataset, denoted by DSEL$^{'}$, is obtained at the end of the process.

\begin{algorithm}[htbp]
	\caption{The Edited Nearest Neighbor rule}
	\label{alg:enn}
	\begin{algorithmic}[1]
		\REQUIRE Dynamic Selection Dataset DSEL
		\STATE DSEL$^{'}$ = DSEL
		\FOR {each $\mathbf{x}_{j,DSEL} \in$ DSEL}
		\IF {$ label\left ( \mathbf{x}_{j,DSEL} \right ) \neq label\left (KNN \left (  \mathbf{x}_{j,DSEL} \right ) \right ) $ }
		\STATE DSEL$^{'}$ = DSEL$^{'} \setminus \left\lbrace  \mathbf{x}_{j,DSEL} \right\rbrace $		
		\ENDIF
		\ENDFOR
		\RETURN DSEL$^{'}$
	\end{algorithmic}
\end{algorithm}

It should be mentioned that the ENN does not remove all samples in the class borders, and that the intrinsic geometry of the class borders and the distribution of the classes are preserved. Only instances that are more likely of being noise, i.e., those for which the majority of neighbors belong to a different class, are removed. Hence, the DS techniques can better estimate the local competence of the base classifiers~\cite{cruz2016prototype}.

\subsection{Relative Neighborhood Graph (RNG)}

The RNG edition technique~\cite{SanchezPF97} is based on the concept of Proximity Graph (PG). In this case, the Relative Neighborhood Graph~\cite{Jaromczyk92relativeneighborhood} is considered as PG. Such graph, $G = (V,E)$, is built by assigning the instances to vertices (V = $DSEL$) and a set of edges $E$, such that $(\mathbf{x}_{i},\mathbf{x}_{j}) \in E$, if and only if they satisfy the given neighborhood criteria (Equation~\ref{eq:rng}):

\begin{equation}
\label{eq:rng}
\begin{matrix}
(\mathbf{x}_{i},\mathbf{x}_{j}) \in E \Leftrightarrow d(\mathbf{x}_{i},\mathbf{x}_{j}) \le max (d(\mathbf{x}_{i},\mathbf{x}_{k}),d(\mathbf{x}_{j},\mathbf{x}_{k}) ) \\ 
\forall\mathbf{x}_{k} \in X, k \neq i, j 
\end{matrix}
\end{equation}

\noindent where $d(\cdot , \cdot) $ is the Euclidean distance between two samples. The RNG is based on the disjoint intersection between two hyperspheres centered at the instances $\mathbf{x}_{i}$ and $\mathbf{x}_{j}$, whose radius is equal to the Euclidean distance between the two instances. Two samples are graph neighbors if and only if this intersection do not contain any other data point from $DSEL$. The graph neighborhood of an instance is the collection of all its graph neighbors. After constructing the graph, the instances whose class differs from the class of the majority of its graph neighbors are removed from $DSEL$.

\section{Experiments}
\label{experiments}

\subsection{Datasets}
\label{sec:datasets}

The experiments were conducted on 30 datasets. Sixteen datasets were taken from the UCI machine learning repository~\cite{Lichman2013}, four from the STATLOG project~\cite{King95statlog}, four from the Knowledge Extraction based on Evolutionary Learning (KEEL) repository~\cite{FdezFLDG11}, four from the Ludmila Kuncheva Collection of real medical data~\cite{lkc}, and two artificial datasets generated with the Matlab PRTOOLS toolbox~\cite{PRTools}. Those 30 datasets were chosen in order to compare the results obtained in this paper with previous results in the literature~\cite{CruzPR,cruz2016prototype,ijcnn2015}.

\begin{table}[htbp]
	\centering
	\caption{Summary of the 30 datasets used in the experiments [Adapted from~\cite{CruzPR}].}
	\label{table:datasets} 
	\resizebox{0.50\textwidth}{!}{
		\begin{tabular}{|l| r| r| r | l| }
			\hline
			\textbf{Database} & \textbf{ No. of Instances} & \textbf{Dimensionality} & \textbf{No. of Classes}   & \textbf{Source} \\
			\hline
			
			\textbf{Adult} & 48842 & 14 & 2 & UCI  \\        
			\hline
			\textbf{Banana}  & 1000 & 2 &	2 &  PRTOOLS  \\
			\hline
			\textbf{Blood transfusion} & 748 & 4 &	2 &  UCI  \\
			\hline 
			\textbf{Breast (WDBC)} & 568 & 30 & 2 &  UCI \\
			\hline
			\textbf{Cardiotocography (CTG)} & 2126 & 21 & 3 &  UCI \\    
			\hline
			\textbf{Ecoli} & 336 & 7 & 8 &  UCI  \\    
			\hline
			\textbf{Steel Plate Faults} & 1941 & 27 & 7 &  UCI \\   
			\hline
			\textbf{Glass} & 214 & 9 & 6  &  UCI  \\                       
			\hline
			\textbf{German credit} & 1000 & 20 &2  &  STATLOG \\
			\hline
			\textbf{Haberman's Survival} & 306 & 3 & 2 & UCI \\
			\hline
			\textbf{Heart} & 270 & 13  & 2  &  STATLOG \\
			\hline
			\textbf{ILPD} & 583 & 10 & 2  &  UCI \\                       
			\hline									
			\textbf{Ionosphere} & 315 &	34 & 2 &  UCI  \\
			\hline
			\textbf{Laryngeal1} & 213 & 16 & 2 &  LKC \\        
			\hline   
			\textbf{Laryngeal3} & 353 & 16 & 3 &  LKC \\    
			\hline
			\textbf{Lithuanian}  & 1000 & 2 & 2 &  PRTOOLS \\
			\hline 
			\textbf{Liver Disorders} & 345 & 6 & 2 & UCI  \\
			\hline
			\textbf{MAGIC Gamma Telescope}  & 19020 & 10 & 2 &  KEEL \\
			\hline  			
			\textbf{Mammographic}  & 961 & 5 & 2 &  KEEL \\
			\hline  
			\textbf{Monk2}  & 4322 & 6 & 2 &  KEEL  \\
			\hline  			
			\textbf{Phoneme} & 5404 & 6 & 2 &  ELENA  \\   
			\hline
			\textbf{Pima} & 768 & 8 & 2 & UCI  \\
			\hline		
			\textbf{Satimage} & 6435 & 19 & 7 & STATLOG \\    
			\hline 			
			\textbf{Sonar} & 208 &	60 & 2 &  UCI \\
			\hline
			\textbf{Thyroid} &  215 & 5 & 3 &  LKC \\
			\hline
			\textbf{Vehicle} & 846 & 18 & 4 &  STATLOG \\
			\hline
			\textbf{Vertebral Column} & 310 & 6 & 2 &  UCI \\          
			\hline	
			\textbf{WDG V1} & 5000 & 21 & 3 &  UCI \\    
			\hline
			\textbf{Weaning} & 302 & 17 & 2 &  LKC \\
			\hline
			\textbf{Wine} & 178 & 13 & 3 &  UCI \\
			\hline
            
		\end{tabular}
	}
\end{table}

\subsection{Experimental protocol}

\begin{table*}[htbp]
	\centering
	\caption{Hyper-parameters used for the PS techniques. The Keel suite version 3.0 was used.}
	\label{tab:parameters}
		\resizebox{1.00\textwidth}{!}{
	\begin{tabular}{|l|l|l|l|l|l|}
		\hline
		\multicolumn{6}{|c|}{Hyper-Parameters}                                                             \\ \hline
		SSMA                                & CHC                                   & GGA  & RMHC & ENN   & RNG  \\ \hline
		1 to 0 Mutation Probability = 0.01  & Population Size = 50                   & Population Size = 51
		  & K = 1 & K = 3    &  Order of the Graph = 1st order
		     \\
		0 to 1 Mutation Probability = 0.001 & Number of Evaluations = 10,000         & 1 to 0 Mutation Probability = 0.01
		  &  Size of S Respect to T = 10     &     &  Type of Selection = Edition
		     \\
		Cross Probability = 1               & Alfa = 0.5          & 0 to 1 Mutation Probability = 0.001 &  Number of Mutations = 10000     &     &  		     \\
		Population Size = 50                & Percentage of Change in Restart = 0.35 & Cross Probability = 0.6 &       &     &       \\
		Number of Evaluations = 10,000       & 0 to 1 Probability in Restart = 0.25  & Number of Evaluations = 10,000 &       &     &       \\
		Alfa = 0.5       & 0 to 1 Probability in Diverge = 0.05   & Alfa = 0.5  &       &     &       \\
		K = 1                               & K = 1                                  &  Number of Neighbours = 1 &       &     &       \\ \hline
	\end{tabular}
}
\end{table*}

For each dataset, the experiments were conducted using 20 replications. For each replication, the datasets were divided using the holdout method~\cite{hastie_09} on the basis of 50\% for training, 25\% for the dynamic selection dataset, $DSEL$, and 25\% for the test set, $\mathcal{G}$. The divisions were performed while maintaining the prior probabilities of each class. For the proposed META-DES-Oracle, 25\% of the training data was used in the meta-training process $\mathcal{T}_{\lambda}$.

For the two-class classification problems, the pool of classifiers was composed of 100 Perceptrons generated using the Bagging technique. For the multi-class problems, the pool of classifiers was composed of 100 multi-class Perceptrons. The use of linear Perceptron classifiers was motivated by the results reported in Section~\ref{sec:synthetic} showing that the META-DES framework can solve non-linear classification problems with complex decision boundaries using only a few linear classifiers. The values of the hyper-parameters, $K$, $K_{p}$ and $h_{c}$, were set at 7, 5 and 70\%, respectively. They were selected empirically based on previous publications~\cite{ijcnn2011,icpr2014,CruzPR}. Hence, the size of the meta-feature vector is 67 ($(7 \times 8)$ $+ 5 + 6$).

The prototype selection techniques were implemented using the KEEL software (version 3.0)~\cite{Alcala-FdezFLDG11}. The hyper-parameters of each PS method were set using the standard values from the KEEL software. They are illustrated in Table~\ref{tab:parameters}. 

\subsection{Results}

In order to know whether the edition of DSEL, by different prototype selection techniques, leads to a significant improvement in classification accuracy by dynamic selection techniques, we conducted a pairwise comparison between the results obtained using the original distribution of DSEL, and those obtained using the six PS techniques analyzed in this paper. We refer to the results obtained using the original distribution of DSEL (without prototype selection) as the baseline result for the rest of this paper.

\begin{figure}[H]
	\begin{center}  	 
		\includegraphics[clip=,  width=0.500\textwidth]{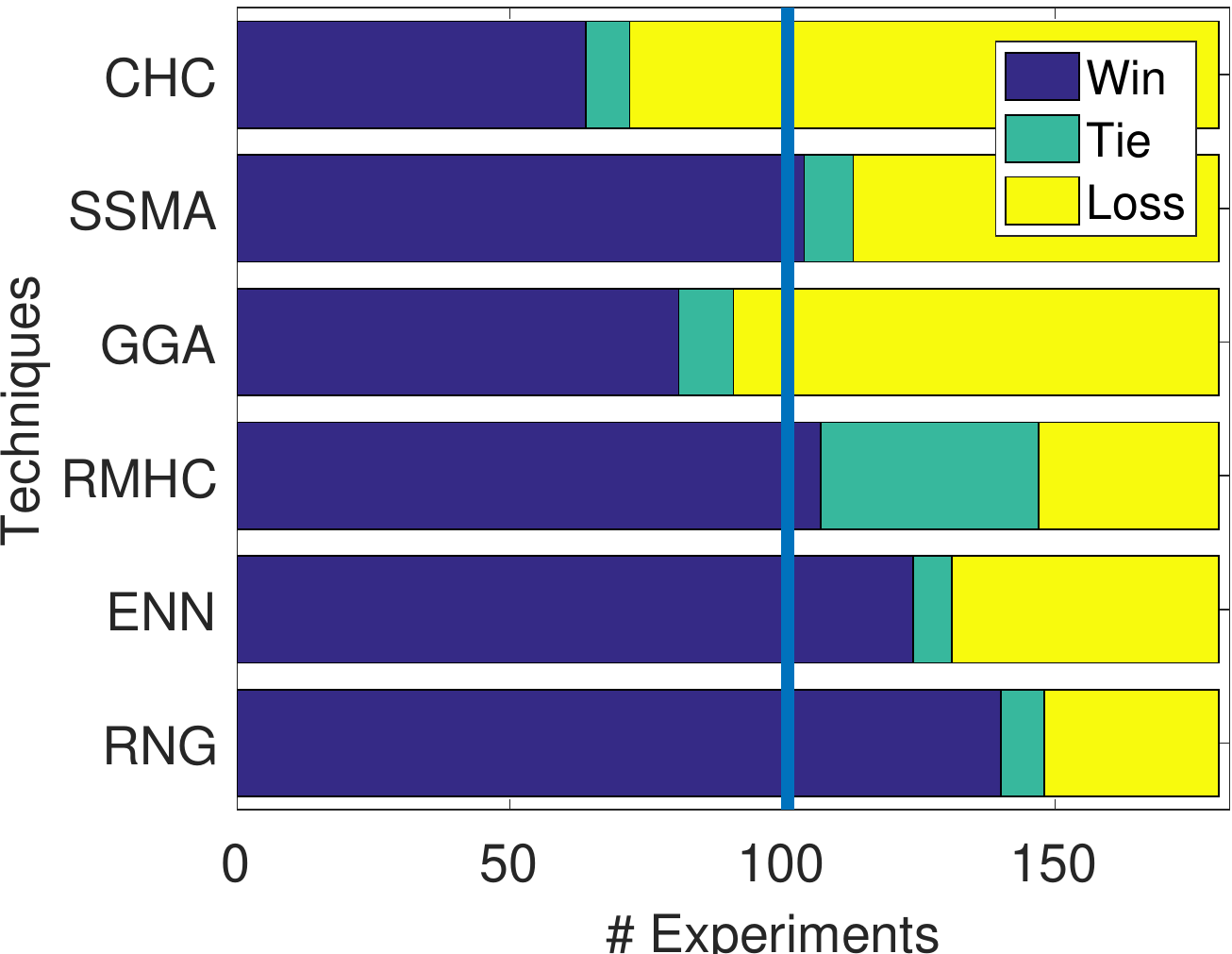}
	\end{center}
	\caption{Pairwise comparison between the results achieved using the PS techniques. The analysis is based in terms of wins, ties and losses. The vertical blue line illustrates the critical value $n_{c} = 101$.}
	\label{fig:wintielossPS}
\end{figure}

 For each PS technique, the six DS methods were evaluated over the 30 datasets, giving a total of 180 experiments (30 datasets $\times$ 6 DS methods). Then, a pairwise analysis was conducted based on Sign test~\cite{Demsar:2006}, computed on the computed wins, ties and losses obtained by each PS technique when compared with the baseline. The null hypothesis, $H_{0}$, meant that both techniques obtained statistically equivalent results. Rejection in $H_{0}$ meant that the classification performance obtained by corresponding prototype selection was significantly better at a predefined significance level $\alpha$. In this paper we set $\alpha = 0.05$ (95\% confidence). The null hypothesis, $H_{0}$, is rejected when the number of wins needs to be greater than or equal to a critical value, denoted by $n_{c}$. The critical value is computed using Equation~\ref{eq:statistical}:

\begin{equation}
\label{eq:statistical}
n_{c} = \frac{n_{exp}}{2} + z_{\alpha}\frac{\sqrt{n_{exp}}}{2}
\end{equation}

\noindent where $n_{exp}$ is the total number of experiments and $z_{\alpha} = 1.645$, for a significance level of $\alpha = 0.05$. For a $n_{exp} = 180$, the critical value is $n_{c} = 101$. Figure~\ref{fig:wintielossPS} shows the results of the Sign test for the six PS techniques. We can observe that all edition techniques (RNG, RMHC and ENN) presented a significant improvement in classification accuracy when compared to the baseline. The RNG method obtained the best performance with 140 wins and 8 ties, followed by the ENN with 124 wins and 7 ties, and the RMHC with 107 wins and 40 ties.

On the other hand, for the hybrid techniques, only the SSMA obtained a significant gain in performance with 104 wins and 9 ties. The two other hybrid techniques presented a higher number of losses than wins (108 losses for the CHC and 89 for the GGA).

To compare the results of all the DES techniques over the 30 classification datasets, the Friedman rank test~\cite{Friedman} was conducted. For each PS technique, the six DS methods were evaluated over the 30 datasets, giving a total of 180 experiments (30 datasets $\times$ 6 DS methods). For each dataset and dynamic selection method, the Friedman test ranks each PS method, with the best performing one getting rank 1, the second best rank 2, and so on. Then, the average rank of each PS was computed. The PS method which obtained the best overall performance is the one presenting the lowest average rank.
 	
\begin{figure*}[htbp]
		\begin{center}  	 
			\includegraphics[clip=,  width=0.60\textwidth]{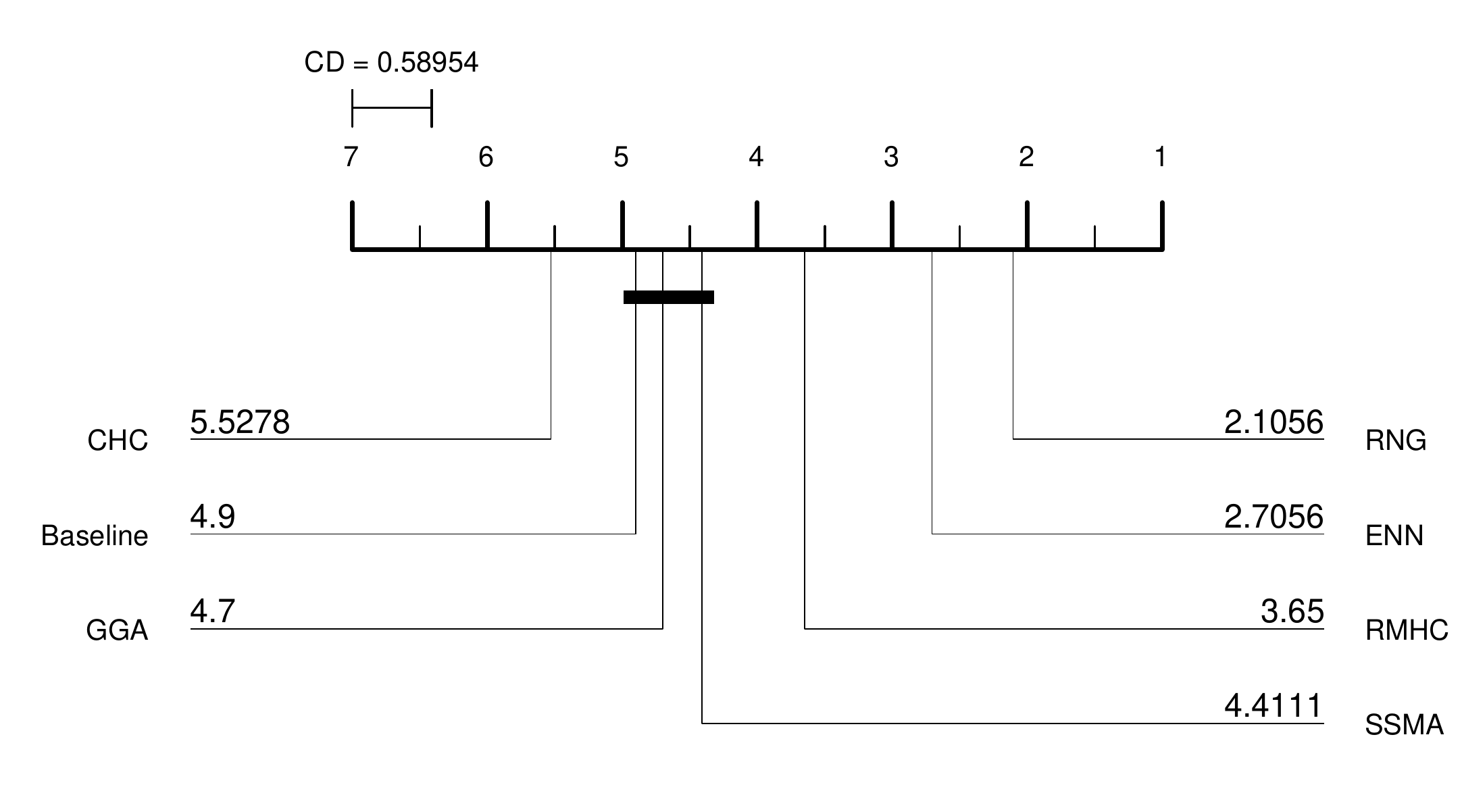}
		\end{center}
		\caption{Critical difference diagram considering the six Prototype Selection techniques. The best algorithm is the one presenting the lowest average rank. Techniques that are statistically equivalent are connected by a black bar. Baseline indicates the result obtained using the original distribution of DSEL.}
		\label{fig:CDdiagram}
	\end{figure*}

Moreover, the post-hoc Bonferroni-Dunn test was applied for a comparison between the ranks achieved by each PS method. The performance of two PS techniques is significantly different if their difference in average rank is higher than the critical difference (CD) calculated using the Bonferroni-Dunn post-hoc test. The average ranks of the six prototype selection techniques, as well as the result of the post-hoc test, are presented using the CD diagram proposed in~\cite{Demsar:2006}. Figure~\ref{fig:CDdiagram} shows the comparison among the six PS techniques. We can see that the three edition techniques considered in this paper, ENN, RNG and RMHC obtained the best overall results (lowest average ranking). Furthermore, their classification results were also statistically superior when compared to the three hybrid techniques, SSMA, GGA and CHC, as well as the original distribution of DSEL.
 
Surprisingly, the CHC method obtained the worst performance in the experiments. Its average rank is significantly worse when compared to the other PS techniques. In addition, even the original distribution of DSEL presented a significantly superior result. Based on the recent survey~\cite{Garcia:2012}, this PS method achieves the highest accuracy for small and medium sized datasets. Therefore, we can also conclude that the best PS technique for improving the classification accuracy of the KNN classifier may not be suitable to improve the performance of dynamic selection techniques. The same phenomenon happened to the GGA technique, which is always among the top 3 techniques in the study conducted in~\cite{Garcia:2012}. However, its result is not significantly superior when compared to those obtained using the original distribution of DSEL.

\begin{figure*}[htbp]
	\centering
	\subfigure[RNG]{\includegraphics[width=2.3in]{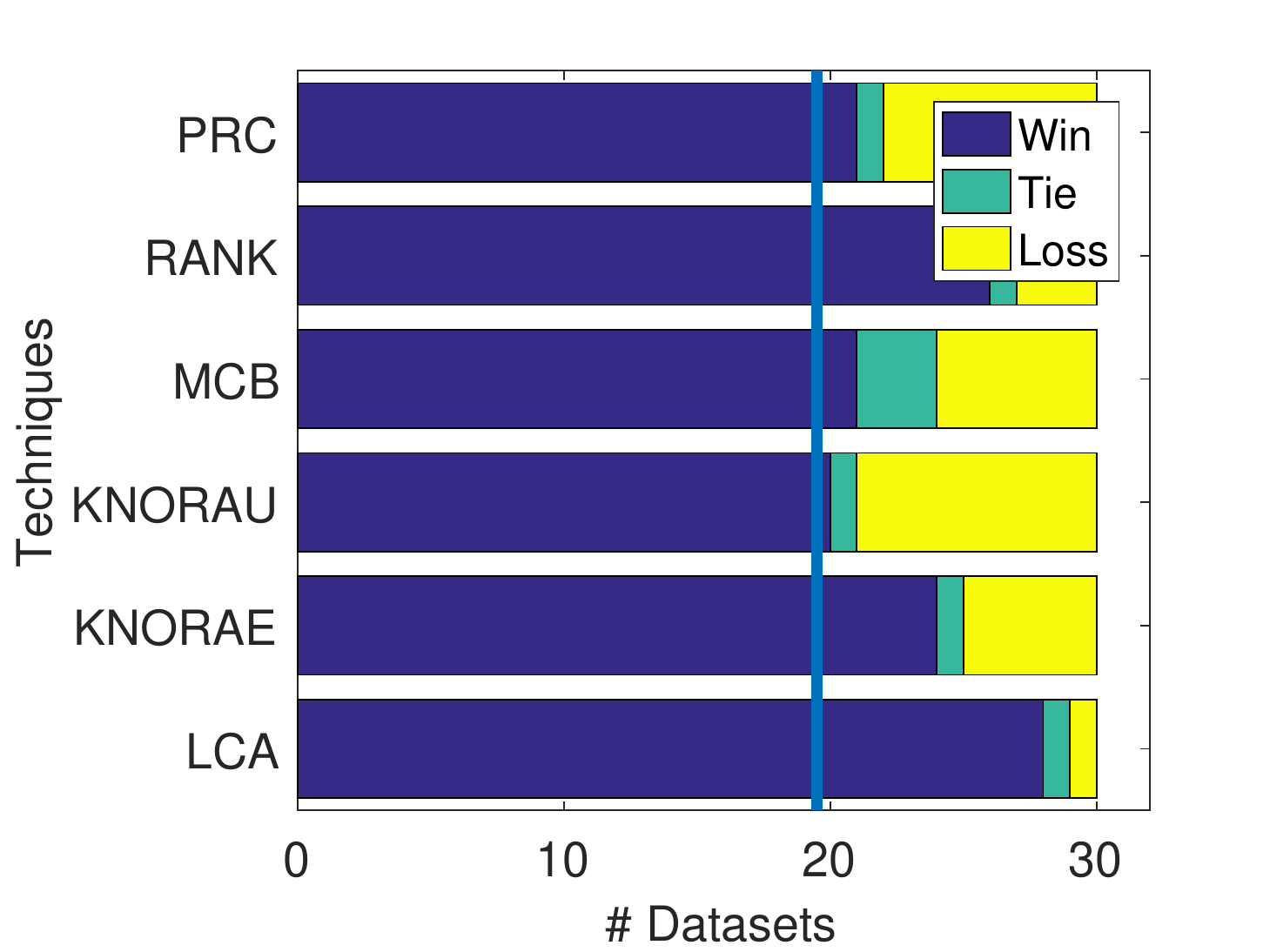}}   
	\subfigure[ENN]{\includegraphics[width=2.3in]{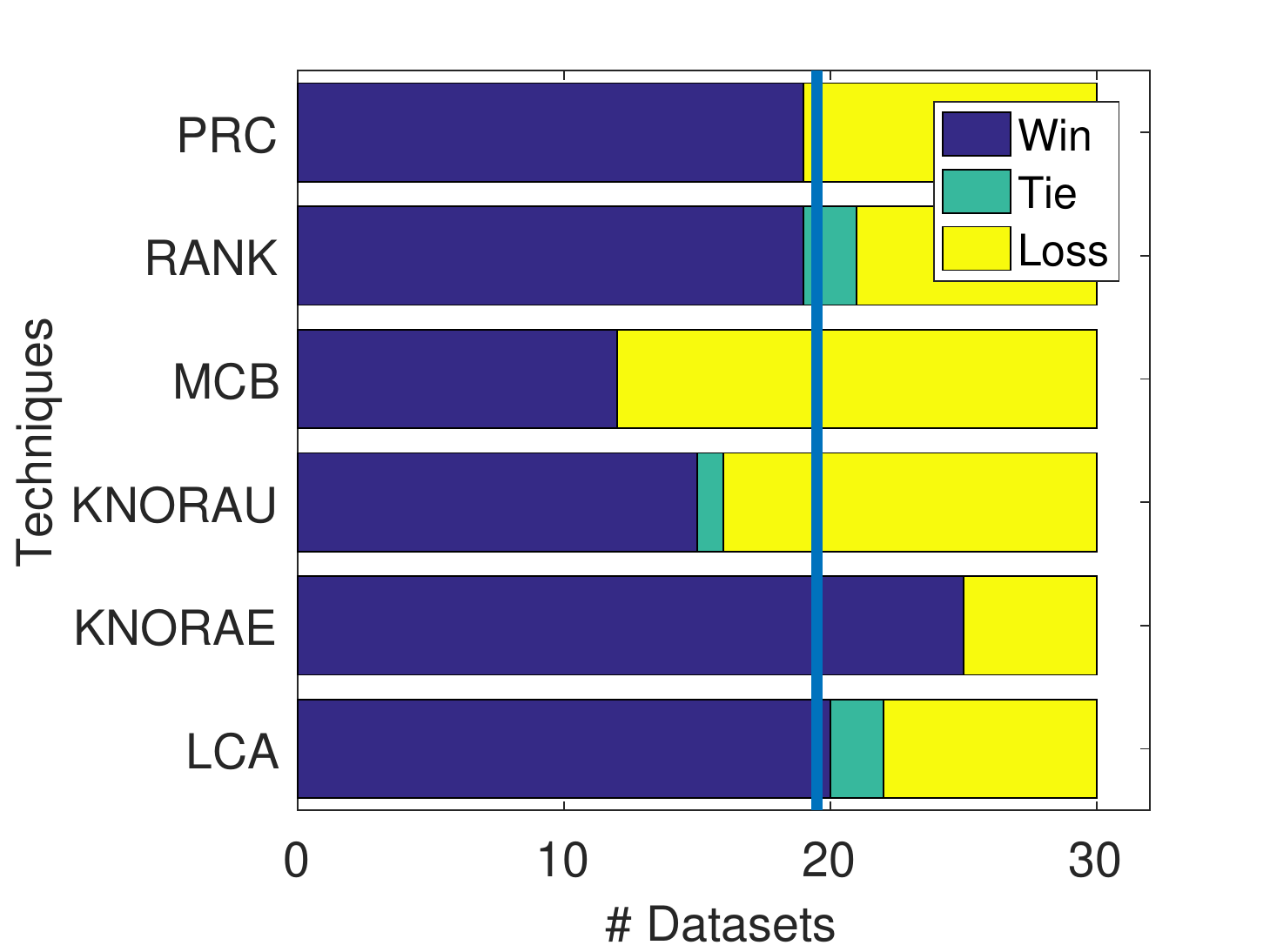}}	
	\subfigure[RMHC]{\includegraphics[width=2.3in]{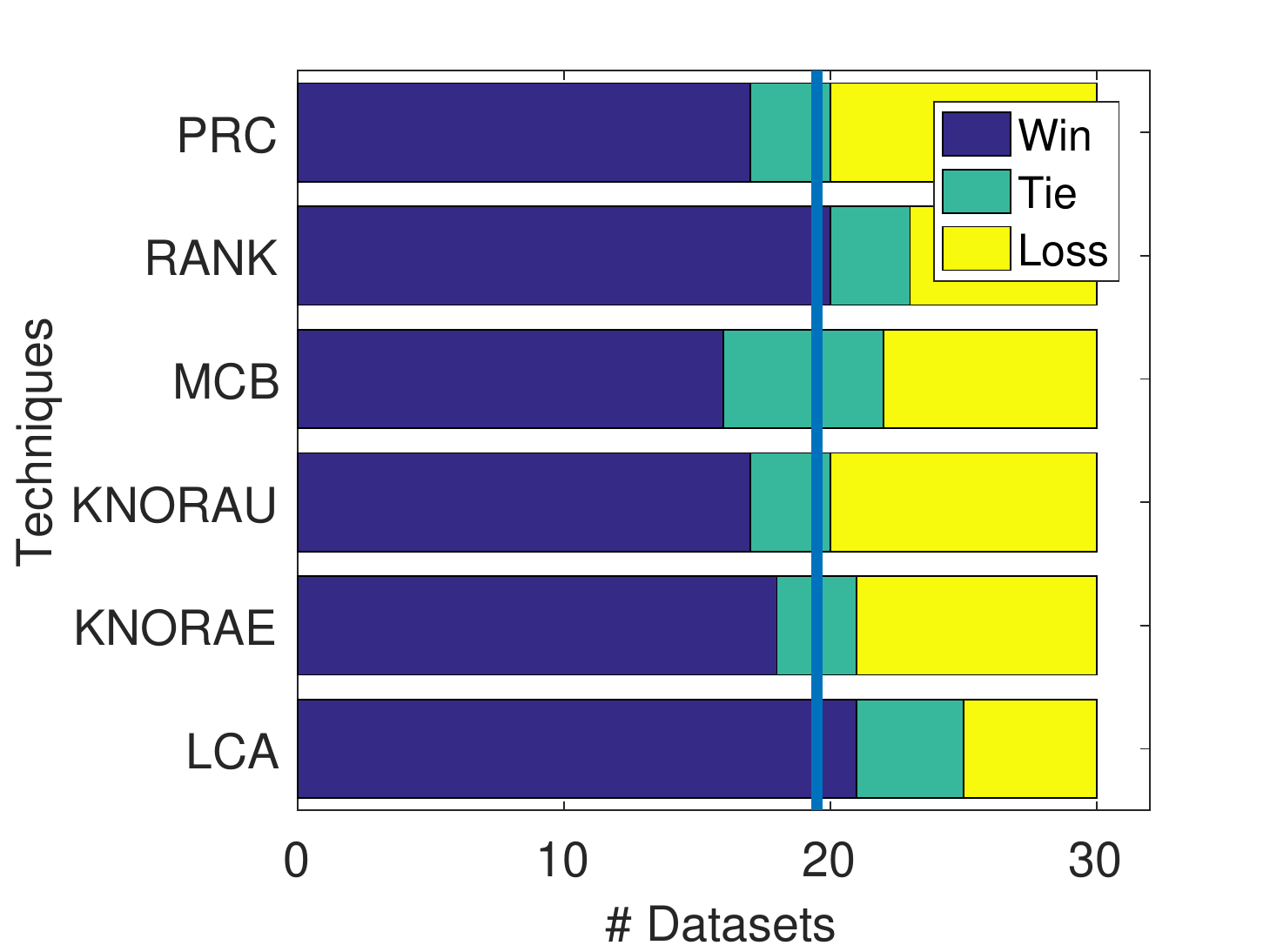}}   
	\subfigure[SSMA]{\includegraphics[width=2.3in]{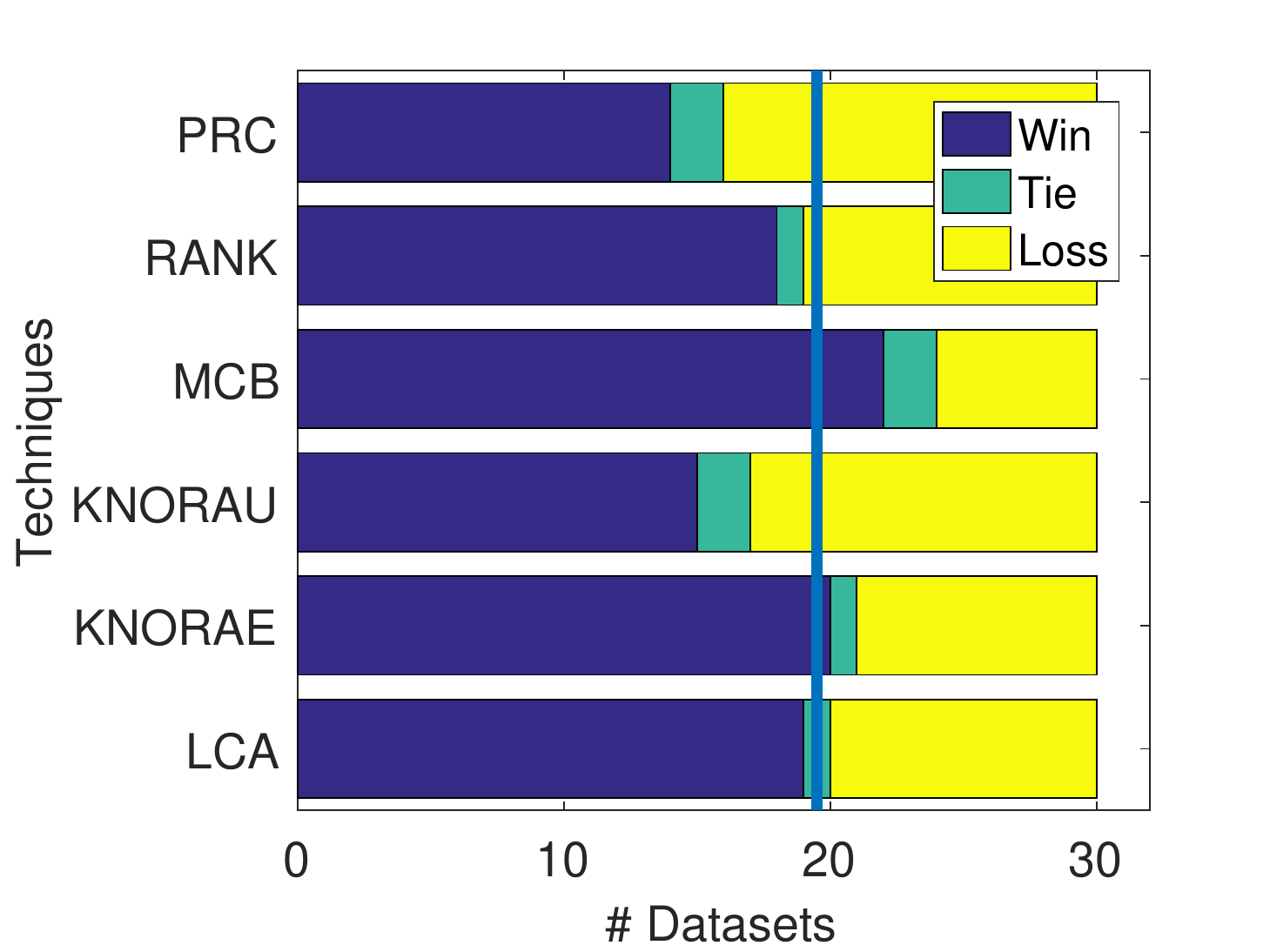}}	
	\subfigure[GGA]{\includegraphics[width=2.3in]{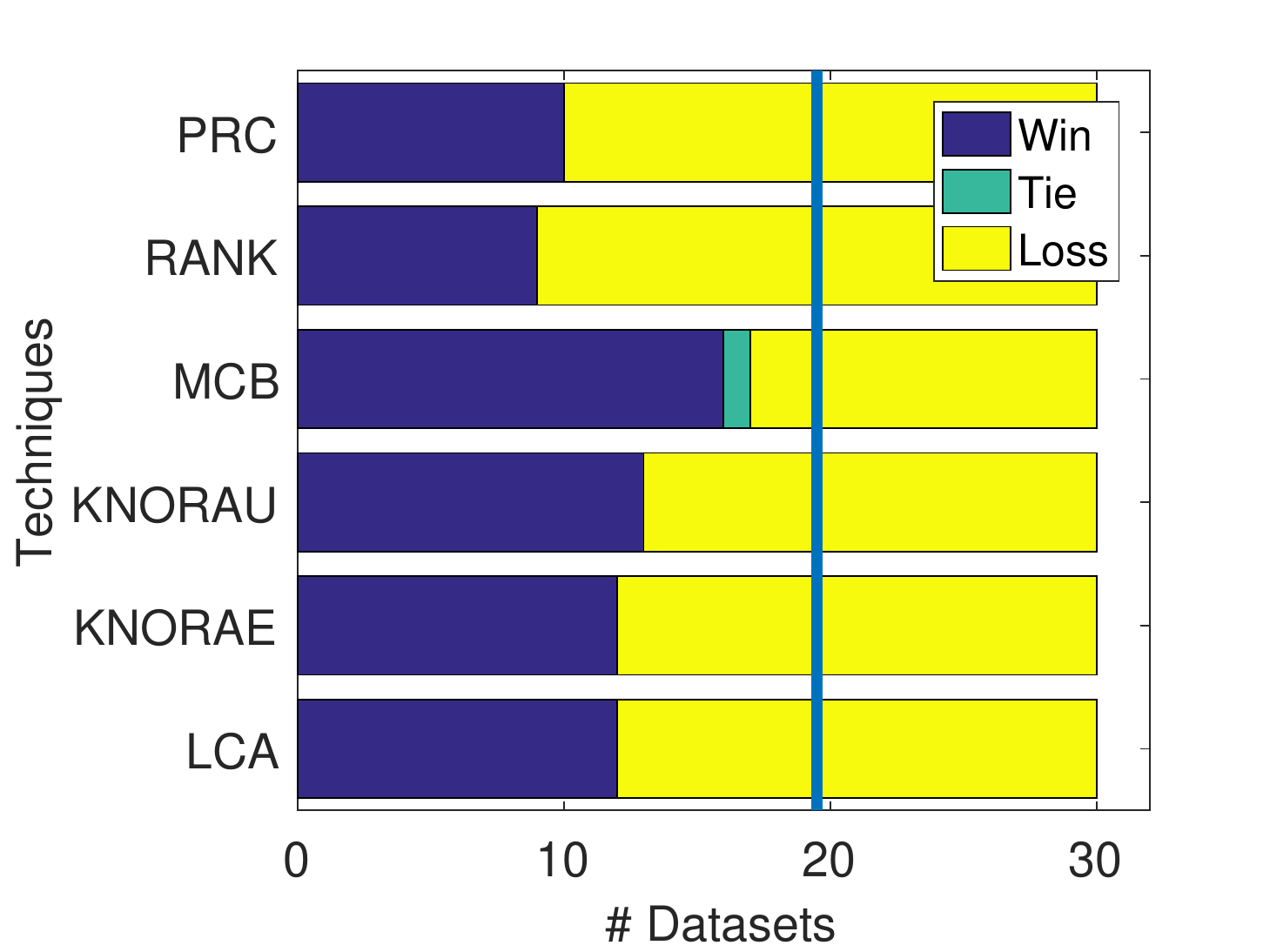}}   
	\subfigure[CHC]{\includegraphics[width=2.3in]{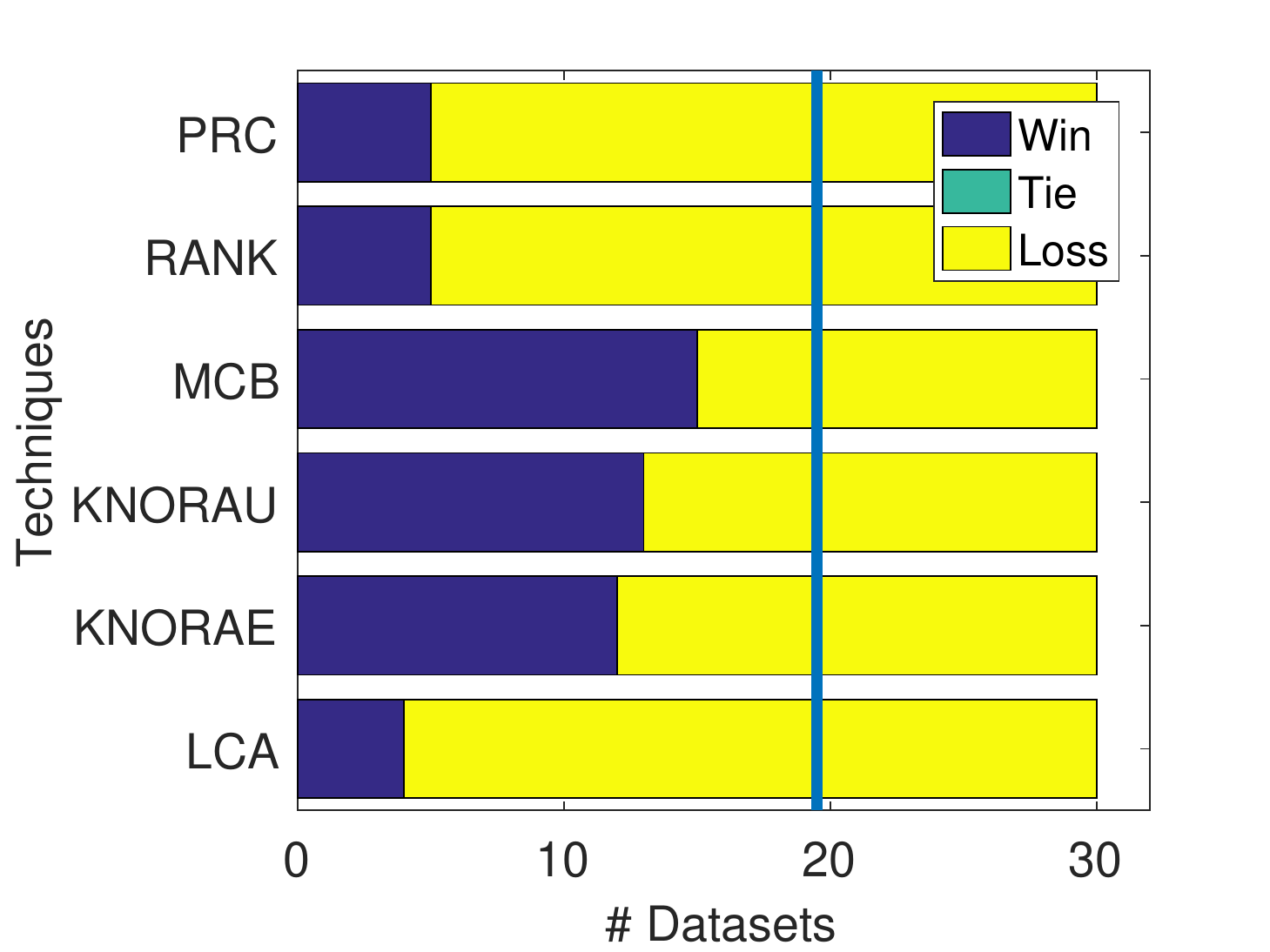}}	
	\caption{Pairwise comparison between the results achieved using the different DS techniques. The analysis is based in terms of wins, ties and losses. The vertical blue line illustrates the critical value $n_{c} = 19.5$ for a confidence level $\alpha = 0.05$, and $n_{exp} = 30$.}
	\label{fig:wintielosDES}
	
\end{figure*}

Furthermore, we also analyze the impact of the prototype selection methods for the six dynamic selection techniques independently. The results of the Sign test are shown in Figure~\ref{fig:wintielosDES}. It can be seen the RNG significantly improved the classification performance of all DS techniques. In contrast, the CHC and GGA reduced the accuracy of all DS techniques.

\subsection{Dataset analysis}
 
\begin{figure}
		\begin{center}  	 
 			\includegraphics[clip=,  width=0.500\textwidth]{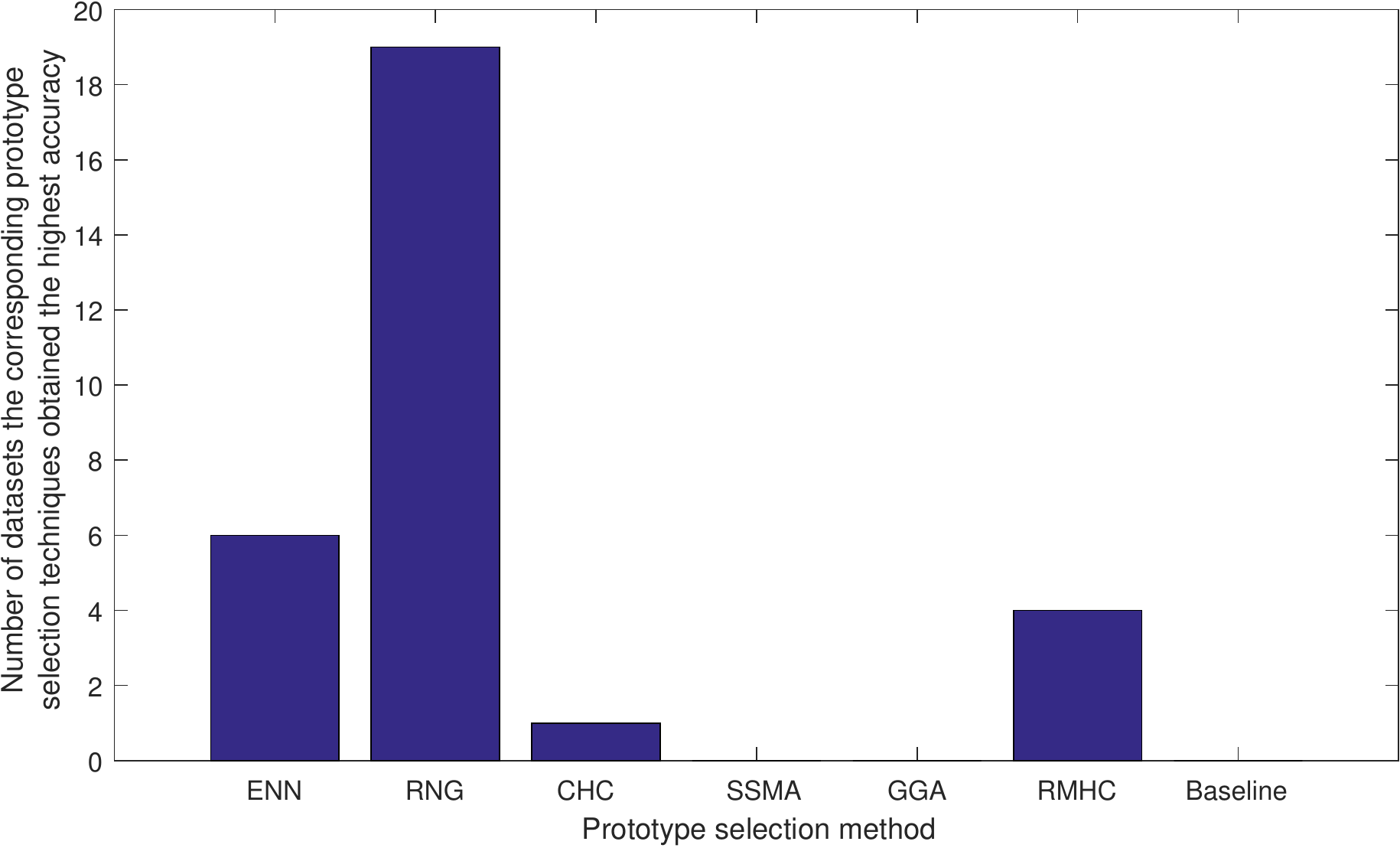}
 		\end{center}
 		\caption{Bar plot showing the number of experiments that each prototype selection technique achieved the highest recognition accuracy.}
 		\label{fig:frequency}
	\end{figure}

In this section, we analyze whether or not different PS techniques are more suitable for different classification problems. Table~\ref{table:datasetResults} shows the average accuracy obtained by the PS techniques for each dataset. The best results are highlighted in bold. Also, Figure~\ref{fig:frequency} shows the number of datasets that each PS method achieved the highest accuracy. For each dataset, the Kruskal-Wallis non-parametric statistical test with a 95\% confidence interval was conducted to perform a pairwise comparison between the baseline and the best results obtained by any of the six PS techniques. Results that are significantly better are marked with a $\bullet$. 

We can observe that the DS using the RNG obtained the highest classification accuracy for the majority of datasets (19 in total), followed by 6 using the ENN and  4 using the RMHC technique. Only for the Mammography dataset, a hybrid technique (SSMA) obtained the highest average accuracy. However, the gain in classification accuracy for this dataset was not statistically significant. Hence, we can conclude that PS methods from the edition paradigm are better suited to improve the classification performance of dynamic selection techniques. Moreover, among the three edition techniques, the one presenting the best result varies (as illustrated by~\ref{fig:frequency}). Thus, the choice of the best PS method may be application dependent.

Another interesting finding comes from the observation that the results using the original DSEL distribution (baseline) never achieved the highest classification accuracy. Hence, the results obtained in this work shows how dynamic selection can be benefited from PS techniques.

\begin{table*}[htbp] 
\centering 
\caption{Comparison of different PS selection techniques. The best results are in bold. Results that are significantly better according to the Kruskal-Wallis non-parametric statistical test with a 95\% confidence interval are marked with a $\bullet$.} 
\label{table:datasetResults}  
\resizebox{0.75\textwidth}{!}{  
	\begin{tabular}{|l | c | c c c c c c|}  
		\hline  

	Dataset & \textbf{Baseline} & \textbf{ENN} & \textbf{RNG} & \textbf{CHC} & \textbf{SSMA} & \textbf{GGA} & \textbf{RMHC} \\ 

	\hline  

	\textbf{Pima} & 74.23(1.76) & 76.36(0.92) & \textbf{76.90(1.19)} $\bullet$ & 72.61(2.14) & 75.18(0.86) & 73.13(2.18) & 76.49(0.88)  \\ 
	\textbf{Liver} & 61.10(3.74) & 65.70(2.81) & \textbf{66.92(3.26)} $\bullet$ & 63.78(1.62) & 64.24(1.48) & 63.29(1.90) & 65.00(1.72)  \\ 
	\textbf{Breast} & 96.73(0.58) & 96.96(0.35) & \textbf{97.01(0.35)} & 96.37(0.25) & 96.16(0.56) & 96.38(0.25) & 96.92(0.34)  \\ 
	\textbf{Blood} & 75.95(1.23) & 76.93(1.04) & \textbf{77.56(1.31)}  & 74.65(0.50) & 75.80(0.48) & 76.05(0.82) & 77.23(1.26)  \\ 
	\textbf{Banana} & 91.44(3.79) & \textbf{94.12(1.95)} $\bullet$ & 93.85(2.20) & 83.15(3.85) & 86.12(3.33) & 83.20(3.95) & 93.56(2.66)  \\ 
	
	\textbf{Vehicle} & 81.37(1.98) & \textbf{82.37(1.57)} & \textbf{82.37(1.17)} & 81.42(1.51) & 81.93(1.48) & 81.66(1.35) & 81.37(1.98)  \\ 
	\textbf{Lithuanian} & 90.53(5.25) & \textbf{91.61(6.24)} & 91.47(6.17) & 81.83(2.28) & 87.01(3.82) & 86.92(3.43) & 91.51(6.16)  \\ 
	\textbf{Sonar} & 77.13(2.65) & 80.77(2.19) & 80.87(2.26) & 76.54(3.32) & 78.24(2.17) & 79.33(2.15) & \textbf{81.19(2.39)} $\bullet$ \\ 
	\textbf{Ionosphere} & 87.52(1.89) & 87.75(0.78) & \textbf{88.22(0.67)} & 85.85(1.97) & 86.00(2.07) & 86.38(2.57) & 87.54(0.78)  \\ 
	\textbf{Wine} & 96.40(2.61) & 97.70(1.19) & \textbf{98.00(0.67)} $\bullet$ & 95.00(5.11) & 95.33(4.89) & 94.96(5.23) & 97.67(1.34)  \\ 
	\textbf{Haberman} & 72.76(1.91) & 75.00(0.50) & 74.80(0.38) & 74.04(1.81) & 73.51(1.36) & 73.25(2.67) & \textbf{75.13(0.30)} $\bullet$ \\ 
	\textbf{CTG} & 85.59(0.84) & 75.00(0.50) & \textbf{86.62(1.32)} & 84.90(0.52) & 85.25(0.86) & 85.40(0.96) & 86.43(1.25)  \\ 
	\textbf{Vertebral} & 84.75(2.59) & 85.02(1.00) & \textbf{85.28(0.69)}  & 84.81(1.54) & 84.76(1.11) & 84.12(1.96) & 84.91(0.94)  \\ 
	\textbf{Faults} & 67.19(0.57) & \textbf{75.02(1.00)} $\bullet$ & 69.95(1.54) & 66.64(1.05) & 67.63(0.63) & 67.71(0.51) & 67.19(0.57)  \\ 
	\textbf{WDVG1} & 83.85(0.69) & 85.02(1.00) & \textbf{85.14(0.52)} $\bullet$ & 84.00(0.55) & 84.15(0.55) & 67.71(0.51) & 83.85(0.69)  \\ 
	\textbf{Ecoli} & 77.28(1.34) & 80.35(2.60) & \textbf{80.53(2.35)} $\bullet$ & 76.80(1.29) & 77.59(1.25) & 76.69(1.91) & 79.82(2.42)  \\ 
	\textbf{GLASS} & 59.16(3.73) & 65.25(3.16) & \textbf{67.20(2.74)} $\bullet$ & 57.89(4.26) & 59.62(3.60) & 59.31(2.91) & 63.68(2.66)  \\ 
	\textbf{ILPD} & 67.72(0.80) & 65.25(3.16) & 67.20(2.74) & 57.89(4.26) & 69.67(0.83) & 59.31(2.91) & \textbf{69.94(1.46)} $\bullet$ \\ 
	\textbf{Adult} & 83.53(2.96) & 87.07(0.45) & \textbf{87.10(0.78)} $\bullet$ & 84.16(3.00) & 85.40(1.14) & 84.63(2.16) & 86.97(0.77)  \\ 
	\textbf{Weaning} & 77.90(2.58) & 80.02(1.94) & \textbf{81.56(2.14)} $\bullet$ & 76.03(2.73) & 78.42(1.30) & 76.23(2.34) & 78.22(2.69)  \\ 
	\textbf{Laryngeal1} & 79.39(1.97) & 81.51(0.41) & 81.07(0.63) & 81.23(0.83) & 80.88(0.52) & 80.66(0.85) & \textbf{82.14(0.59)} $\bullet$ \\ 
	\textbf{Thyroid} & 96.38(0.37) & \textbf{96.82(0.16)} & 96.78(0.16) & 96.46(0.55) & 96.81(0.33) & 96.60(0.36) & 96.58(0.19)  \\ 
	\textbf{Laryngeal3} & 70.33(2.23) & 72.62(0.86) & \textbf{72.75(0.46)} $\bullet$ & 68.30(6.07) & 70.56(2.74) & 69.51(3.89) & 72.23(0.95)  \\ 
	\textbf{German} & 72.60(1.92) & 75.09(0.70) & \textbf{75.49(0.79)} $\bullet$ & 72.68(2.61) & 73.41(1.41) & 73.90(1.21) & 75.13(0.43)  \\ 
	\textbf{Heart} & 82.18(1.86) & 84.83(1.55) & \textbf{85.39(1.85)} $\bullet$ & 81.18(2.42) & 82.30(1.74) & 81.37(2.63) & 84.39(1.29)  \\ 
	\textbf{Segmentation} & 95.17(0.64) & 95.57(0.26) & \textbf{95.75(0.29)} & 95.01(0.68) & 95.15(0.49) & 95.20(0.45) & 95.17(0.64)  \\ 
	\textbf{Phoneme} & 77.51(3.36) & 95.57(0.26) & \textbf{95.75(0.29)} & 95.01(0.68) & 79.05(4.29) & 95.20(0.45) & 77.51(3.36)  \\ 
	\textbf{Monk2} & 81.33(2.87) & 88.97(5.68) & \textbf{89.43(5.93)} $\bullet$ & 79.32(1.52) & 84.46(2.45) & 84.26(2.51) & 86.42(4.20)  \\ 
	\textbf{Mammographic} & 82.11(1.57) & 83.41(0.82) & 83.51(0.51) & \textbf{83.80(0.37)} & 83.29(1.10) & 83.00(1.33) & 83.09(1.01)  \\ 
	\textbf{Magic} & 79.40(2.55) & \textbf{82.40(1.25)} $\bullet$ & 81.30(1.25) & 80.20(2.05) & 80.10(1.85) & 80.67(1.63) &  81.67(1.63)  \\ 
		\hline 
	\end{tabular}} 
\end{table*}

\subsection{Computational complexity analysis}
\label{sec:compression}

	\begin{table}[htbp]
		\centering
		\caption{Reduction rate obtained by each PS technique.}
		\label{table:reduction} 
		\resizebox{0.45\textwidth}{!}{
			\begin{tabular}{|c| c| c|}  
				\hline
				\textbf{PS Method}  & \textbf{dataset red.(\%)}  & \textbf{red. comp. time(\%)}  \\ 
				\hline
				\textbf{ENN}  & 21.26 &  16.66 \\ 
				\hline	
				\textbf{RNG}   &  20.09 & 17.56 \\ 
				\hline
				\textbf{RMHC}  &  60.19 &  30.12 \\ 
				\hline
				\textbf{SSMA}  &  90.55 &   68.76 \\ 
				\hline
				\textbf{CHC}  & 98.08  &   73.60 \\ 
				\hline
				\textbf{GGA}  & 96.54 &   73.25 \\ 
				\hline

			\end{tabular}
		}
	\end{table}

In this section, we analyze the trade-off between dataset reduction and accuracy obtained by the PS techniques. Table~\ref{table:reduction} shows the average reduction rate obtained by each PS technique. We can see that the reduction rate of the edition techniques is much lower when compared to the hybrid ones. In addition, the techniques which presented the worst dataset reduction rates (RNG and ENN) presented the best overall performance. 

In contrast, the techniques which presented the highest reduction rates were the ones with the worst classification accuracy (GGA and CHC). For some datasets such as Sonar and Heart, only three instances were kept in DSEL. As reported in~\cite{reportarXiv}, DS techniques may not perform well, when there is not enough sample in DSEL. Hence, we believe that techniques presenting a very high reduction rate may not be suitable to be used with DS. 

The results obtained using the SSMA technique are promising since it presented a very high reduction rate, on average more than 95\% of the instances were removed, as well as good classification performance when compared to the baseline (original distribution of DSEL). The SSMA significantly reduced the computational time involved to run the DS techniques. Thus, by using the SSMA, it is possible significantly reduce the computational cost involved during the generalization steps in dynamic selection, while having a slightly better performance when compared to the baseline (as shown in Figures~\ref{fig:wintielossPS} and~\ref{fig:CDdiagram}). Even though this technique is behind the edition ones in terms of recognition rate, it presented a good trade-off between classification accuracy and computational time. 

\section{Discussion and Conclusion} 
\label{conclusion}

In this paper, we evaluate six prototype selection techniques in order to edit the distribution of the dynamic selection dataset, DSEL. The analysis is conducted using six state-of-the-art dynamic selection techniques and is based on the classification accuracy and computational complexity points of view.

Based on the experimental study, all three PS techniques from the edition paradigm presented a significant improvement in classification accuracy when used to edit DSEL. Moreover, the RNG method achieved better results when compared to the previously proposed ENN. For the Hybrid PS techniques, only the SSMA presented a slight improvement in classification accuracy. The CHC and GGA techniques reduced the performance of the system. Furthermore, the edition methods outperformed the three hybrid ones. Thus, based on this analysis, edition PS techniques are the better candidate to be used in conjunction with dynamic selection. 

The only hybrid technique that showed an improvement in classification performance for the DS methods was the SSMA. This technique was able to significantly reduce the size of DSEL (average compression rate of 95\%), and achieve good classification performance. This result is promising since by reducing the size of DSEL the computational time required to apply the DS methods is significantly reduced. An interesting direction for future work would involve the use the performance of DS techniques, (e.g., the accuracy of the  LCA technique) as a criterion inside those techniques in order to determine which data points should be removed from DSEL, i.e., use the classification accuracy of a DS technique, such as, the LCA rather than the accuracy of the 1NN to evaluate the fitness of each dataset generated.

\section*{Acknowledgment}

This work was supported by the Natural Sciences and Engineering Research Council of Canada (NSERC), the \'{E}cole de technologie sup\'{e}rieure (\'{E}TS Montr\'{e}al), CNPq (Conselho Nacional de Desenvolvimento Cient\'{i}fico e Tecnol\'{o}gico) and FACEPE (Funda\c{c}\~{a}o de Amparo \`{a} Ci\^{e}ncia e Tecnologia de Pernambuco).
 
\bibliographystyle{IEEEtran}
\bibliography{report}

 \newcommand{\noop}[1]{}
\begin{thebibliography}{10}
\providecommand{\url}[1]{#1}
\csname url@samestyle\endcsname
\providecommand{\newblock}{\relax}
\providecommand{\bibinfo}[2]{#2}
\providecommand{\BIBentrySTDinterwordspacing}{\spaceskip=0pt\relax}
\providecommand{\BIBentryALTinterwordstretchfactor}{4}
\providecommand{\BIBentryALTinterwordspacing}{\spaceskip=\fontdimen2\font plus
\BIBentryALTinterwordstretchfactor\fontdimen3\font minus
  \fontdimen4\font\relax}
\providecommand{\BIBforeignlanguage}[2]{{%
\expandafter\ifx\csname l@#1\endcsname\relax
\typeout{** WARNING: IEEEtran.bst: No hyphenation pattern has been}%
\typeout{** loaded for the language `#1'. Using the pattern for}%
\typeout{** the default language instead.}%
\else
\language=\csname l@#1\endcsname
\fi
#2}}
\providecommand{\BIBdecl}{\relax}
\BIBdecl

\bibitem{kittler}
J.~Kittler, M.~Hatef, R.~P.~W. Duin, and J.~Matas, ``On combining
  classifiers,'' \emph{IEEE Transactions on Pattern Analysis and Machine
  Intelligence}, vol.~20, pp. 226--239, 1998.

\bibitem{kuncheva}
L.~I. Kuncheva, \emph{Combining Pattern Classifiers: Methods and
  Algorithms}.\hskip 1em plus 0.5em minus 0.4em\relax Wiley-Interscience, 2004.

\bibitem{Alceu2014}
A.~S. Britto, R.~Sabourin, and L.~E.~S. de~Oliveira, ``Dynamic selection of
  classifiers - {A} comprehensive review,'' \emph{Pattern Recognition},
  vol.~47, no.~11, pp. 3665--3680, 2014.

\bibitem{CruzPR}
R.~M.~O. Cruz, R.~Sabourin, G.~D.~C. Cavalcanti, and T.~I. Ren, ``{META-DES:}
  {A} dynamic ensemble selection framework using meta-learning,'' \emph{Pattern
  Recognition}, vol.~48, no.~5, pp. 1925--1935, 2015.

\bibitem{knora}
A.~H. Ko, R.~Sabourin, and A.~S. Britto~Jr, ``From dynamic classifier selection
  to dynamic ensemble selection,'' \emph{Pattern Recognition}, vol.~41, pp.
  1735--1748, May 2008.

\bibitem{paulo2}
P.~R. Cavalin, R.~Sabourin, and C.~Y. Suen, ``Dynamic selection approaches for
  multiple classifier systems,'' \emph{Neural Computing and Applications},
  vol.~22, no. 3-4, pp. 673--688, 2013.

\bibitem{lca}
K.~Woods, W.~P. Kegelmeyer, Jr., and K.~Bowyer, ``Combination of multiple
  classifiers using local accuracy estimates,'' \emph{IEEE Transactions on
  Pattern Analysis Machine Intelligence}, vol.~19, pp. 405--410, April 1997.

\bibitem{mcb}
G.~Giacinto and F.~Roli, ``Dynamic classifier selection based on multiple
  classifier behaviour,'' \emph{Pattern Recognition}, vol.~34, pp. 1879--1881,
  2001.

\bibitem{Woloszynski}
T.~Woloszynski and M.~Kurzynski, ``A probabilistic model of classifier
  competence for dynamic ensemble selection,'' \emph{Pattern Recognition},
  vol.~44, pp. 2656--2668, October 2011.

\bibitem{WoloszynskiKPS12}
T.~Woloszynski, M.~Kurzynski, P.~Podsiadlo, and G.~W. Stachowiak, ``A measure
  of competence based on random classification for dynamic ensemble
  selection,'' \emph{Information Fusion}, vol.~13, no.~3, pp. 207--213, 2012.

\bibitem{reportarXiv}
R.~M.~O. Cruz, R.~Sabourin, and G.~D.~C. Cavalcanti, ``A {DEEP} analysis of the
  {META-DES} framework for dynamic selection of ensemble of classifiers,''
  \emph{CoRR}, vol. abs/1509.00825, 2015.

\bibitem{cruz2016prototype}
R.~M. Cruz, R.~Sabourin, and G.~D. Cavalcanti, ``Prototype selection for
  dynamic classifier and ensemble selection,'' \emph{Neural Computing and
  Applications}, pp. 1--11, 2016.

\bibitem{ijcnn2011}
R.~M. O.~Cruz, G.~D. C.~Cavalcanti, and T.~I. Ren, ``A method for dynamic
  ensemble selection based on a filter and an adaptive distance to improve the
  quality of the regions of competence,'' \emph{Proceedings of the
  International Joint Conference on Neural Networks}, pp. 1126 -- 1133, 2011.

\bibitem{enn}
D.~L. Wilson, ``Asymptotic properties of nearest neighbor rules using edited
  data,'' \emph{IEEE Transactions on Systems, Man and Cybernetics}, vol.~2,
  no.~3, pp. 408--421, July 1972.

\bibitem{T1976}
I.~Tomek, ``An experiment with the edited nearest-neighbor rule,'' \emph{IEEE
  Transactions on Systems and Man and Cybernetics}, vol.~6, no.~6, pp.
  448--452, 1976.

\bibitem{Garcia:2012}
S.~Garcia, J.~Derrac, J.~Cano, and F.~Herrera, ``Prototype selection for
  nearest neighbor classification: Taxonomy and empirical study,'' \emph{IEEE
  Transactions on Pattern Analysis and Machine Intelligence}, vol.~34, no.~3,
  pp. 417--435, 2012.

\bibitem{Skalak94}
D.~B. Skalak, ``Prototype and feature selection by sampling and random mutation
  hill climbing algorithms,'' in \emph{Machine Learning, Proceedings of the
  Eleventh International Conference, Rutgers University, New Brunswick, NJ,
  USA, July 10-13, 1994}, 1994, pp. 293--301.

\bibitem{SanchezPF97}
J.~S. S{\'{a}}nchez, F.~Pla, and F.~J. Ferri, ``Prototype selection for the
  nearest neighbour rule through proximity graphs,'' \emph{Pattern Recognition
  Letters}, vol.~18, no.~6, pp. 507--513, 1997.

\bibitem{Garcia:2008}
S.~Garc\'{\i}a, J.~R. Cano, and F.~Herrera, ``A memetic algorithm for
  evolutionary prototype selection: A scaling up approach,'' \emph{Pattern
  Recognition}, vol.~41, no.~8, pp. 2693--2709, Aug. 2008.

\bibitem{Eshelman90}
L.~J. Eshelman, ``The {CHC} adaptive search algorithm: How to have safe search
  when engaging in nontraditional genetic recombination,'' in \emph{Proceedings
  of the First Workshop on Foundations of Genetic Algorithms. Bloomington
  Campus, Indiana, USA, July 15-18 1990.}, 1990, pp. 265--283.

\bibitem{KunchevaJ99}
L.~Kuncheva and L.~C. Jain, ``Nearest neighbor classifier: Simultaneous editing
  and feature selection,'' \emph{Pattern Recognition Letters}, vol.~20, no.
  11-13, pp. 1149--1156, 1999.

\bibitem{Kuncheva95}
L.~I. Kuncheva, ``Editing for the k-nearest neighbors rule by a genetic
  algorithm,'' \emph{Pattern Recognition Letters}, vol.~16, no.~8, pp.
  809--814, 1995.

\bibitem{Jaromczyk92relativeneighborhood}
J.~W. Jaromczyk and G.~T. Toussaint, ``Relative neighborhood graphs and their
  relatives,'' in \emph{Proc. IEEE}, 1992, pp. 1502--1517.

\bibitem{Lichman2013}
\BIBentryALTinterwordspacing
K.~Bache and M.~Lichman, ``{UCI} machine learning repository,'' 2013. [Online].
  Available: \url{http://archive.ics.uci.edu/ml}
\BIBentrySTDinterwordspacing

\bibitem{King95statlog}
R.~D. King, C.~Feng, and A.~Sutherland, ``Statlog: Comparison of classification
  algorithms on large real-world problems,'' 1995.

\bibitem{FdezFLDG11}
J.~Alcal{\'{a}}{-}Fdez, A.~Fern{\'{a}}ndez, J.~Luengo, J.~Derrac, and
  S.~Garc{\'{\i}}a, ``{KEEL} data-mining software tool: Data set repository,
  integration of algorithms and experimental analysis framework,''
  \emph{Multiple-Valued Logic and Soft Computing}, vol.~17, no. 2-3, pp.
  255--287, 2011.

\bibitem{lkc}
\BIBentryALTinterwordspacing
L.~Kuncheva, ``Ludmila kuncheva collection {LKC},'' 2004. [Online]. Available:
  \url{http://pages.bangor.ac.uk/~mas00a/activities/real_data.htm}
\BIBentrySTDinterwordspacing

\bibitem{PRTools}
\BIBentryALTinterwordspacing
R.~P.~W. Duin, P.~Juszczak, D.~de~Ridder, P.~Paclik, E.~Pekalska, and D.~M.
  Tax, ``Prtools, a matlab toolbox for pattern recognition,'' 2004. [Online].
  Available: \url{http://www.prtools.org}
\BIBentrySTDinterwordspacing

\bibitem{ijcnn2015}
R.~M.~O. Cruz, R.~Sabourin, and G.~D.~C. Cavalcanti, ``{META-DES.H:} {A}
  dynamic ensemble selection technique using meta-learning and a dynamic
  weighting approach,'' in \emph{International Joint Conference on Neural
  Networks (IJCNN}, 2015, pp. 1--8.

\bibitem{icpr2014}
------, ``On meta-learning for dynamic ensemble selection,'' in \emph{22nd
  International Conference on Pattern Recognition, {ICPR} 2014, Stockholm,
  Sweden, August 24-28, 2014}, 2014, pp. 1230--1235.

\bibitem{Alcala-FdezFLDG11}
J.~Alcal{\'{a}}{-}Fdez, A.~Fern{\'{a}}ndez, J.~Luengo, J.~Derrac, and
  S.~Garc{\'{\i}}a, ``{KEEL} data-mining software tool: Data set repository,
  integration of algorithms and experimental analysis framework,''
  \emph{Multiple-Valued Logic and Soft Computing}, vol.~17, no. 2-3, pp.
  255--287, 2011.

\bibitem{Demsar:2006}
J.~Dem\v{s}ar, ``Statistical comparisons of classifiers over multiple data
  sets,'' \emph{Journal of Machine Learning Research}, vol.~7, pp. 1--30, Dec.
  2006.

\bibitem{Friedman}
J.~H. Friedman and L.~C. Rafsky, \emph{The Annals of Statistics}, vol.~7,
  no.~4, pp. 697--717, 1979.

\end{thebibliography}

\end{document}